\documentclass[11pt,a4paper]{article}
\usepackage[hyperref]{emnlp-ijcnlp-2019}
\usepackage{times}
\usepackage{latexsym}
\usepackage{graphicx}
\usepackage{algorithm, algorithmic, amsthm}
\usepackage{amssymb}
\usepackage{amsmath}
\usepackage{multirow}
\usepackage{subfigure}
\usepackage{url}

\aclfinalcopy

\title{Task-Oriented Conversation Generation Using\\Heterogeneous Memory Networks}

\author{Zehao Lin $^1$ \qquad Xinjing Huang $^1$ \qquad Feng Ji $^2$ \qquad Haiqing Chen $^2$ \qquad Yin Zhang$^1$\thanks{\textsuperscript{*}Corresponding Author}  \\ 
  $^1$ College of Computer Science and Technology, Zhejiang University\\
  $^2$ DAMO Academy, Alibaba Group \smallskip \\
  \texttt{\{georgelin,huangxinjing,zhangyin98\}@zju.edu.cn}\\ \texttt{\{zhongxiu.jf,haiqing.chenhq\}@alibaba-inc.com}
}

\date{}

\begin{document}
\maketitle
\begin{abstract}
How to incorporate external knowledge into a neural dialogue model is critically important for dialogue systems to behave like real humans. To handle this problem, memory networks are usually a great choice and a promising way. However, existing memory networks do not perform well when leveraging heterogeneous information from different sources. In this paper, we propose a novel and versatile external memory networks called Heterogeneous Memory Networks (HMNs), to simultaneously utilize user utterances, dialogue history and background knowledge tuples. In our method, historical sequential dialogues are encoded and stored into the context-aware memory enhanced by gating mechanism while grounding knowledge tuples are encoded and stored into the context-free memory. During decoding, the decoder augmented with HMNs recurrently selects each word in one response utterance from these two memories and a general vocabulary. Experimental results on multiple real-world datasets show that HMNs significantly outperform the state-of-the-art data-driven task-oriented dialogue models in most domains.
\end{abstract}

\section{Introduction}
Compared with chitchat, task-oriented dialogue systems aim at solving tasks in specific domains with grounding knowledge. Though far from handling conversation like a real human, existing task-oriented dialogue systems have shown cheerful prospect in a specific domain, e.g. Siri and Cortana are personal assistants, helping people a lot in daily life and business work. 

\begin{figure}
\includegraphics[width=1.0\linewidth]{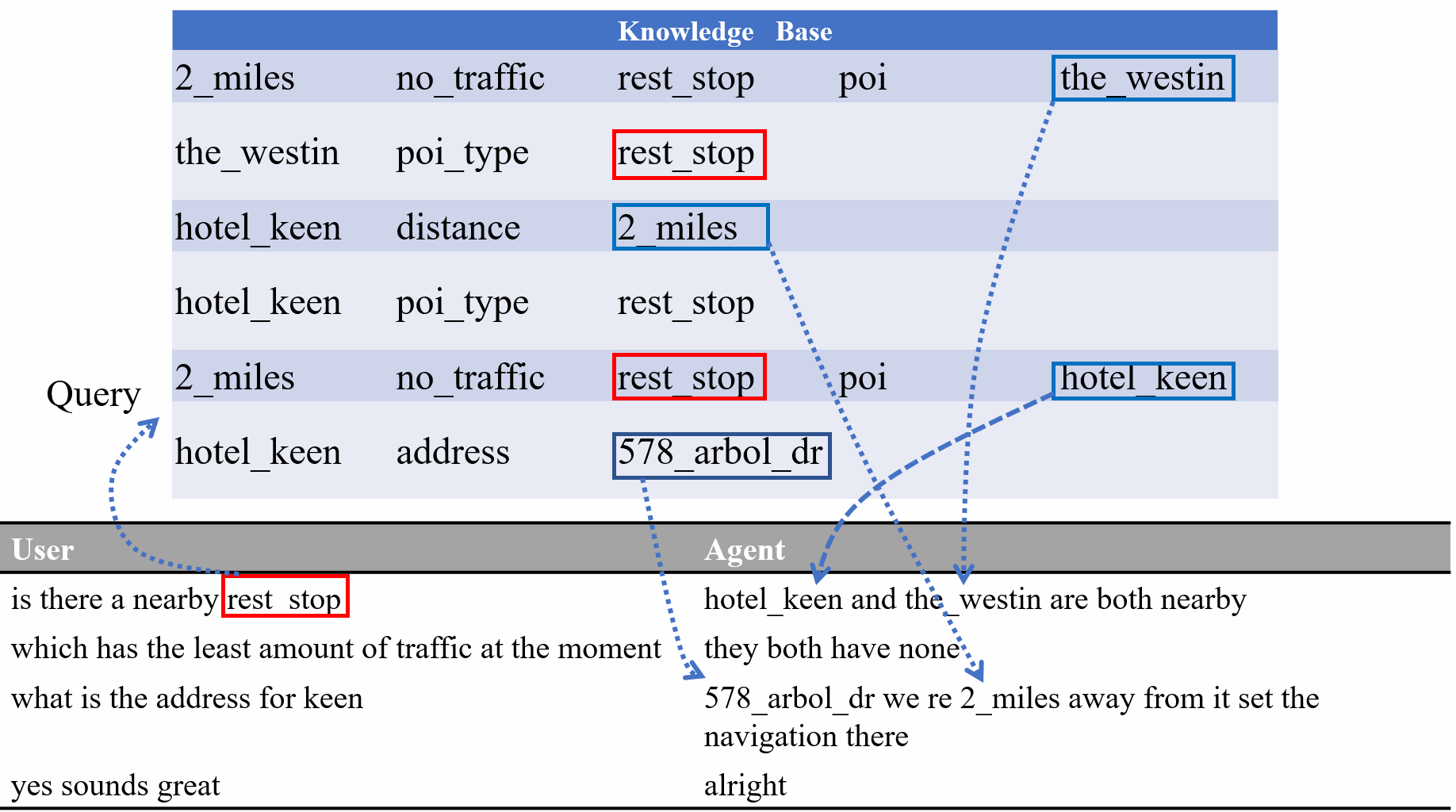}
\caption{A multi-turn dialogue example. The upper table shows several n-tuples sampled from knowledge base. Lower table shows multi-turn dialogues. Agent needs to retrieve appropriate knowledge tuples to generate the proper response.}
\label{fig:example_digloa}
\end{figure}

In general, knowledge-grounded task-oriented dialogue system can be divided into three important components: understanding user utterances, fetching right knowledge from external storage and replying right answer. As shown in Figure \ref{fig:example_digloa}, agent is required to do a point-of-interest navigation. According to dialogue history, agent will fetch related knowledge base information, in our case represented as tuples (e.g. [$hotel\_keen$, $poi\_type$, $rest\_stop$], which indicates the point-of-i nterests type of $hotel\ keen$ is $rest\ stop$), as an external knowledge to answer correctly and complete task.

Traditional pipeline dialogue systems\cite{YanDCZZL17,wen2016network}  and some end-to-end dialogue systems rely on the predefined the slot filling labels. Despite the consumption of human efforts, these kinds of systems are difficult to adapt to new domains. 

Many works, e.g. \cite{vinyals2015neural,DBLP:conf/acl/ShangLL15},  show that training a fully data-driven end-to-end model is a promising way to build domain-agnostic dialogue system. Their models mostly try to use the attention mechanism, including memory networks techniques, to fetch the most similar knowledge \cite{DBLP:conf/nips/SukhbaatarSWF15}, then incorporate grounding knowledge into a seq2seq neural model to generate a suitable response \cite{DBLP:conf/acl/FungWM18}. 

However, existing memory networks equally treat information from multiple sources, e.g. sequential dialogue history and structure knowledge bases. Therefore two weaknesses arise in such methods: (1) It is difficult to model different types of structured information in only one memory network. (2) It is also difficult to model the effectiveness of knowledge from different sources in such a single memory network. To address these issues, we expand the architecture of memory networks used in a seq2seq neural model. 

Our contributions are mainly three-fold:
\begin{itemize}
    \item We propose a novel seq2seq neural conversation model augmented with Heterogeneous Memory Networks. We first model sequential dialogue history and grounding external knowledge with two different kinds of memory networks and then feed the output of context-aware memory to the context-free memory to search the representations of similar knowledge.
    \item Our context-aware memory networks is able to learn the context-aware latent representations and stores them into memory slots, by employing a gating mechanism when encoding dialogue history and user utterance.
    \item Experimental results 
    demonstrate that our neural approach significantly outperforms the examined neural methods automatic metrics, and context-aware memory networks can learn and store more meaningful representations than the examined memory approaches.
\end{itemize}

\begin{figure*}
\includegraphics[width=1.0\linewidth]{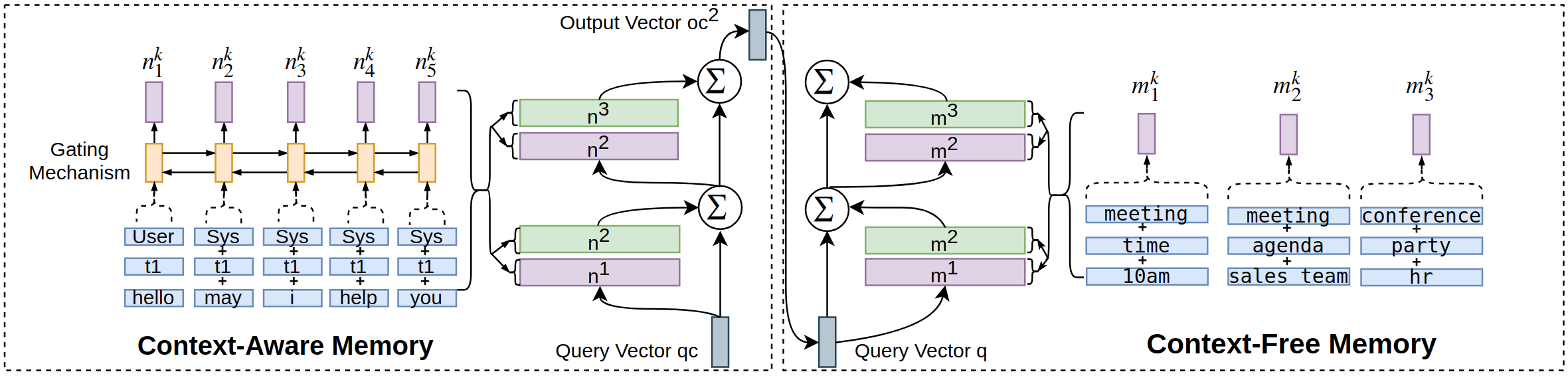}
\caption{An example of Heterogeneous Memory Networks with two-hop attention. Context-aware memory which encodes dialog history to a context vector $oc^2$ while context-free memory loads knowledge base information. The output of context-aware memory will be employed as the query vector to the context-free memory.}
\label{fig:model_overview}
\end{figure*}

\section{Related Works}
The end-to-end model uses deep neural net instead of several parts in pipeline models to generate responses. \cite{wen2016network} propose a data-driven goal-oriented neural dialogue system by adding database operator and policy networks modules to introduce database information and track state which need extra labeling step that breaks differentiability. \cite{bordes2016learning} propose a testbed to break down the strengths and shortcomings of end-to-end dialog systems in goal-oriented applications. Those methods treated dialogue system as the problem of learning a mapping policy from dialogue histories to agents' responses. 

The booming internet dialogue data lay the foundation of building data-driven models. \cite{DBLP:conf/emnlp/RitterCD11} first applied phrase-based Statistical Machine Translation\cite{DBLP:conf/ijcnlp/SetiawanLZO05}. It treats the conversation system as a translation problem, a user utterance needs to be translated into an agent response. 

\cite{NIPS2014_5346} propose Sequence to sequence model (SEQ2SEQ) architecture and apply it to neural machine translation task. SEQ2SEQ has become a general basis of natural language generation tasks, e.g. question answering\cite{DBLP:conf/aaai/TanWYDLZ18} and question generation \cite{DBLP:conf/nlpcc/ZhouYWTBZ17}. By applying the RNN based encoder-decoder framework to generate responses, models\cite{DBLP:conf/acl/ShangLL15, DBLP:conf/emnlp/ChoMGBBSB14, DBLP:conf/acl/LuongSLVZ15} are able to utilize neural networks to learn the representation of dialogue histories and generate appropriate responses.

To deal with multi-turn information, \cite{DBLP:conf/naacl/SordoniGABJMNGD15} propose a model that represents the whole dialogue history (including the current message) with continuous representations or embeddings of words and phrases to address the challenge of the context-sensitive response generation.

By adding a knowledge base module, recent works \cite{ghazvininejad2017knowledge,young2018augmenting} have shown the possibility of training an end-to-end task-oriented dialogue system on the sequence to sequence architecture. Ghazvininejad et al. \cite{ghazvininejad2017knowledge} generalize the SEQ2SEQ approach by conditioning responses on both conversation history and external knowledge, aiming at producing more contextual responses without slot filling. 

CopyNet \cite{DBLP:conf/acl/GuLLL16} and Pointer Networks \cite{NIPS2015_5866} improve model's accuracy and ability of handling of out-of-vocabulary words using neural attention. Pointer-Generator networks \cite{DBLP:conf/acl/SeeLM17} apply copy mechanism to the neural generation model. Their work shows copy mechanism can improve quality in text generation. \cite{dhingra2016towards} and \cite{li2017end} apply reinforcement learning to make it differentiable. 

Recent works on external memory \cite{graves2014neural,DBLP:journals/corr/HenaffWSBL16} provide an efficient method of introducing and reasoning different types of external information.
\cite{DBLP:conf/nips/SukhbaatarSWF15} propose end-to-end memory networks with multiple attention hops model over a possibly large external memory. 
\cite{DBLP:conf/acl/FungWM18} propose Mem2Seq that combines the end-to-end memory networks with the idea of pointer networks. \cite{DBLP:conf/www/ChenRTZY18} add the hierarchical structure and the variational memory network to capture both the high-level
abstract variations and long-term memories during the dialogue tracking. To take care of information from different sources, \cite{DBLP:conf/naacl/FuF18} propose an attention mechanism to encourage the decoder to actively interact with the memory by taking its heterogeneity into account.

\section{Proposed Framework}
To generate responses using dialogue history and grounding knowledge, we introduce a novel encoder-decoder neural conversation model augmented with Heterogeneous Memory Networks (HMNs). The encoder module adopts a context-aware memory network to better understand the dialogue history and query. The decoder is enhanced with HMNs, which is able to incorporate external knowledge and dialog history when generating words.

\begin{figure}
\begin{center}
\includegraphics[width=1.0\linewidth]{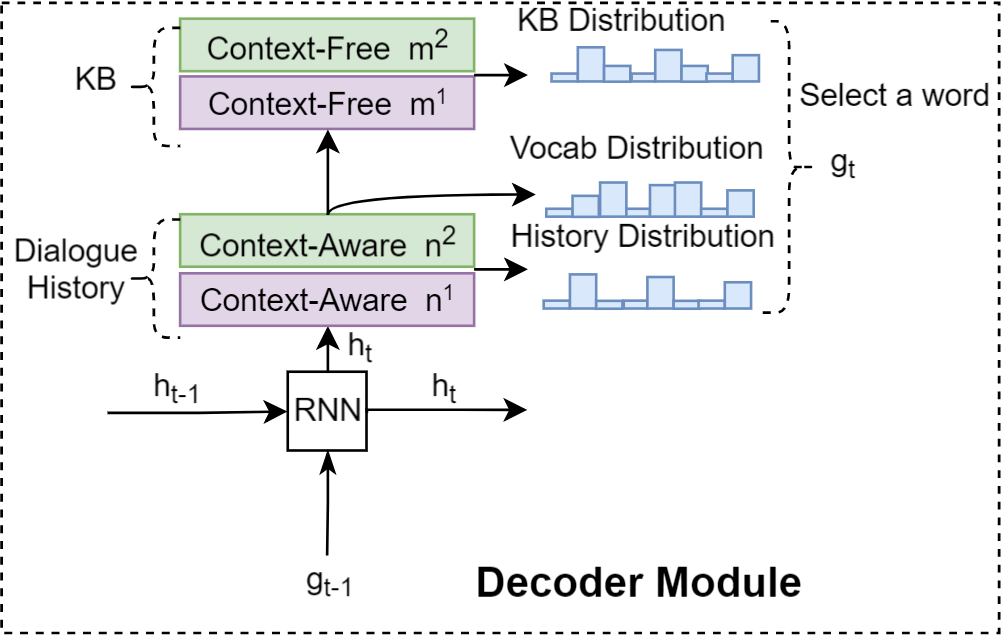}
\caption{Decoder module with two attention hops}
\label{fig:decoder_model}
\end{center}
\end{figure}

\subsection{Encoder} \label{sec:encoder}
The encoder encodes the dialogue history into a fixed context vector. Here we adopt the context-aware memory as our encoder module. As shown in the left part of Figure \ref{fig:model_overview}, each word will be extended to the following parts: 1) token itself. 2) A turn tag. 3) An identity tag. For example, in the first turn a user says "hello" and the response from system is "may I help you", it will be concatenated as [(hello, t1, user),(may, t1, sys),(I, t1, sys),(help, t1, sys),
(you, t1, sys)], \textit{sys} means the word comes from the dialog system, so does \textit{user}. \textit{t1} indicates the word is from the first turn. Each word can be transformed into vector by embedding lookup, and we sum up vectors in each tuple to be the input sequence of the context-aware memory. Using a fixed vector to query the memory, a context vector $c$ can be obtained. 

\subsubsection{Context-Aware Memory}\label{CA-Section}
To efficiently model the context information of sequential data, we present the context-aware memory. Memory slots $n^k$ are structured by concatenating all input vectors as $n^k=cat[n^k_1, n^k_2,..., n^k_l]$, k stands for the k-th hop in memory, and $l$ means the length of the input. $n^k_l$ is the sum of each word tag embedding using each hop own randomly initialized embedding matrix $C^k$.
And we adopt the adjacent weight sharing scheme, which means $C^k$ is not only the input embedding matrix in the k-th hop, but also the output embedding matrix in the (k-1)-th hop.
We add a gating mechanism between memory cells. The gating mechanism applied is adopted from Bidirectional GRU \cite{DBLP:conf/ssst/ChoMBB14} in our case. Thus the context-dependent representation of inputs, denoted as $n^k=[\overrightarrow{n^k},\overleftarrow{n^k}]$, where $\overrightarrow{n^k}$ and $\overleftarrow{n^k}$ are the forward and backward representation of inputs, respectively. The forward process can be illustrated as equations:
    \begin{equation}
         r_t = \sigma(W_1 v^k_t +  W_2 \overrightarrow{n^k}_{(t-1)} + b_1) \label{GRU_starts}
    \end{equation}
    \begin{equation}
         z_t = \sigma(W_3 v^k_t +  W_4 \overrightarrow{n^k}_{(t-1)} + b_2)
    \end{equation}
    \begin{equation}
        e_t = \tanh(W_5 v^k_t + r_t (W_6 \overrightarrow{n^k}_{(t-1)}+ b_3))
    \end{equation}
    \begin{equation}
        \overrightarrow{n^k_t} = (1 - z_t) e_t + z_t \overrightarrow{n^k}_{(t-1)} \label{GRU_ends}
    \end{equation}
where $W_1$, $ W_2$, $W_3$, $W_4$, $W_5$, $W_6$ and $b_1$, $b_2$, $b_3$ are trainable weights and biases. 

Given the query vector ${qc}^k$, attention weights over memory cells $n^k$ can be calculated by the equation:
    \begin{equation}
        p^k= Softmax({n}^k \cdot {{qc}^k})      \label{memory_starts}
    \end{equation}
the readout vector is the sum of output memory matrix ${n}^{k+1}$ with corresponding attention weights $p^k$.
    \begin{equation}
        u^k = \sum_i(p^k_i \cdot {n}^{k+1}_i)
    \end{equation}
    
By summing query and readout vector together,  we can get the output from the k-th hop.
\begin{equation}
	{oc}^k = {qc}^k+u^k  \label{memory_ends}
\end{equation}
 Note ${oc}^k$ is also the query vector of the (k+1)-th hop.

\subsection{Decoder} \label{sec:decoder}
The decoder contains HMNs and an RNN controller, as shown in Figure \ref{fig:decoder_model}. 
The controller controls the process of querying HMNs. 

In each time step, HMNs will generate: 1) a readout vector ${oc}^1$, which is the output of the first hop in history memory, and 2) attention weights of the last hop in two memories, called history word distribution $P_{his}$ and knowledge tuple distribution $P_{kb}$. The readout vector is concatenated with $h_t$ to predict the vocabulary distribution $P_{vocab}$. Formally,
\begin{equation}
    P_{vocab} = {Softmax}(W_7[h_t,{oc}^1]) 
\end{equation}
where $W_7$ is trainable weight matrix. \\
We adopt a simple strategy ( Section \ref{sec:word_select} ) to select a word from three distributions $P_{vocab}$, $P_{his}$ and $P_{kb}$.

\subsubsection{Heterogeneous Memory Networks}
HMNs stacks two types of memory: 1) context-aware memory and 2) context-free memory. Dialog history is loaded into context-aware memory, and knowledge base triples are loaded into context-free memory. Firstly, HMNs accepts query vector as inputs, then walk through context-aware memories. The final output $u^k$ in the last hop will be employed to query context-free memory. Context-aware memory has been detailed in Section \ref{CA-Section}.

Context-Free memory itself is end-to-end memory networks \cite{DBLP:conf/nips/SukhbaatarSWF15}. Compared with our context-aware memory, it has no gating mechanism. The input to the memory is the summed vectors in each knowledge triple, as depicted in the right part of Figure \ref{fig:model_overview}. Each hop owns randomly initialized embedding matrix ${C^k}^{'}$  We denote memory slots as $m^k$. It accepts a query vector and then follows the same process \ref{memory_starts} to \ref{memory_ends}, the output ${u^k}^{'}$ can be obtained.

\subsubsection{Controller}
We adopt GRU as our controller. It accepts the output $c$ from encoder as initial hidden state $h_0$. In each time step, it takes the previous generated word $g_{t-1}$'s embedding $E(g_{t-1})$ and last time hidden state $h_{t-1}$ as inputs. Formally:
    \begin{equation}
        h_t = GRU(E(g_{t-1}),h_{t-1}) 
    \end{equation}
then $h_t$ is used to query the HMNs. 

\subsection{Copy Mechanism}
We adopt copy mechanism to copy words from memories. 
Attention weights in the last hop of the two memories,$P_{kb}$ and $P_{his}$  will be the probability of the target word from those memories. If the target word does not appear in inputs, the position index will be the last position in memories, which is a sentinel added in preprocessing stage. 

\subsection{Joint Learning}
To learn the distribution of three vocabularies $P_{vocab}$, $P_{kb}$ 
and $P_{his}$ in each time step, the loss in the t-th time step is the negative log-likelihood of the predict probability of the target word for that time step. Formally:
\begin{equation}
    Loss = - \frac{1}{T}\sum_{t=0}^{t=T}\sum_i(\log p_{ti})
\end{equation}
Note that $p_{it}$ means the t-th word's probability 
in $i\in{\{P_{vocab}, P_{kb},P_{his}\}}$ . 
    
\begin{table*}[h]
	\centering
		\caption{The statistics of the bAbI-3, 4, 5, DSTC2 and Key-Value Retrieval datasets}
		\label{tab:data_stastics}
		\vspace{-0.1in}
		\begin{tabular}{ l|l | l |l | l| l }
			\hline \hline
			Datasets      & Key-Value Retrieval dataset & DSTC 2 &bAbI-3 &bAbI-4 &bAbI-5 \\
			\hline
			Avg. History words & 25.5 & 63.4 & 49.9& 20.5 & 62.6\\
			Avg. KB pairs      & 64.7 & 42.7 &  23.4 & 7.0 & 23.6 \\
			Avg. Response Length  & 8.7 & 10.2 & 7.2 & 5.7 & 6.5 \\
			Vocabulary Size  & 1554 &1066 & 739 & 1004 & 1135 \\
			Dialogue Turns     & 2.8 & 9.9 & 10.9 & 4.5 & 19.3 \\
			\hline \hline
		\end{tabular}
\end{table*}

\subsection{Word Selection Strategy}\label{sec:word_select}
In our case, if words with the highest probability in $P_{his}$ and $P_{kb}$ vocabularies are not on sentinel positions, we directly compare the probability of each word and select the higher one. If one of the vocabularies points to the sentinel position, the model will select the word with the highest probability in the other vocabulary. At last, if both vocabularies get to sentinel positions, the word from $P_{vocab}$ will be selected.

\section{Experimental Setup}
\subsection{Datasets} \label{sec:datasets}
As the proposed approach is quite general, the model can be applied to any task-oriented dialogue datasets with conversation and knowledge base data. To evaluate and compare the results with the state-of-the-art methods in multiple dimensions, we choose three popular task conversation datasets including DSTC 2, Key-Value Retrieval dataset and the (6) dialog bAbI tasks. Table \ref{tab:data_stastics} shows the statistics of datasets.

\begin{itemize}
    \item \emph{Key-Value Retrieval dataset} \cite{DBLP:conf/sigdial/EricKCM17}. This dataset releases a corpus of 3,031 multi-turn dialogues. The dialogues consist of three different domains: calendar scheduling, weather information retrieval, and point-of-interest navigation.
    \item \emph{The (6) dialog bAbI tasks} \cite{bordes2016learning}. The (6) dialog bAbI tasks are a set of five subtasks within the goal-oriented context of restaurant reservations. Conversations are grounded with an underlying knowledge base of restaurants and their properties (location, type of cuisine, etc.). As task 1 and 2 have been achieved very well, we only test our model on task 3 to 5 and their OOV(out-of-vocabulary), where entities (e.g. restaurant names) in test sets may not have been able to see during training. 
    \item \emph{The Dialog State Tracking Challenge 2} (DSTC 2). DSTC 2 is a research challenge focused on improving the state-of-the-art in tracking the state of spoken dialogue systems. DSTC 2's training dialogues were gathered using Amazon Mechanical Turk related to restaurant search. 
\end{itemize}
\begin{table*}[h]

\centering
\caption{Results on Key-Value Retrieval dataset. F1 score, including Entity F1, is micro-average over the entire set, and three subtasks. Human results are reported by Eric et al.\cite{DBLP:conf/sigdial/EricKCM17}}
\label{table:results_kvr}
\begin{tabular}{ l|l|l|l| l|l }
\hline \hline
Model name  &    BLEU & Ent. F1  & Scheduling Ent. F1 & Weather Ent. F1 & Navigation Ent. F1\\
\hline
Human*    & 13.5 & 60.7 & 64.3 & 61.6 & 55.2 \\  
\hline
SEQ2SEQ    & 11.07 &  30.5   & 30.7 & \textbf{46.4}  & 13.4 \\
SEQ2SEQ+Attn.    & 11.19 & 35.6 & 40.5 & 44.0 & 23.0 \\ 
Mem2Seq   & 12.06 &  31.1 & 51.8 & 34.3 & 12.3 \\    
\hline
HMNs       & \textbf{14.46}  & \textbf{43.1}  & \textbf{61.3} & 40.3 & \textbf{32.3} \\
\hline \hline
\end{tabular}
\end{table*}
\begin{table*}[h]
\centering
\caption{Results of Per-response accuracy and Per-dialog accuracy (in brackets) on bAbI dialogues. Per-dialog accuracy presents the accuracy of complete dialogues.}
\label{table:results_babi}
\vspace{-0.1in}
\begin{tabular}{ c|c c |c c c}
\hline \hline
Task  &    SEQ2SEQ  & SEQ2SEQ+Attn.  & Mem2Seq  & HMNs-CFO & HMNs \\
\hline
T3   &      74.8(0) & 74.8(0) &  83.9(15.6)    &  93.7(55.9) & \textbf{93.6(56.1)}      \\
T4   &      56.5(0) & 56.5(0) &  97.0(90.5)    & 96.8(89.3) & \textbf{100(100)}     \\
T5   &      98.9(82.9) & \textbf{98.6(83)}  &  96.2(46.4)   & 97.1(58.2)  & 98.0(69.0) \\
T3-OOV  &   74.9(0) & 74.0(0)&  83.6(18.1)   & 92.3(45.2)  &  \textbf{92.5(48.2)}\\
T4-OOV  &   56.5(0) & 57.0(0)&  97.0(89.4)   & 96.1(90.3) & \textbf{100(100)} \\
T5-OOV  &   67.2(0) & 67.6(0)&  71.4(0)    & 78.3(0)  & \textbf{84.1(2.6)}   \\
\hline \hline
\end{tabular}
\end{table*}
\begin{table}[h]
\centering
\caption{The results on the DSTC 2}
\label{table:results_dstc2}
\vspace{-0.1in}
\begin{tabular}{ l|p{2cm}|p{2cm}}
\hline \hline
Model name  &    F1 & BLEU  \\

\hline
SEQ2SEQ    & 69.7 & 55.0 \\
SEQ2SEQ+Attn. & 67.1 & \textbf{56.6} \\
SEQ2SEQ+Copy  & 71.6  & 55.4  \\
\hline
Mem2Seq  &   75.3 &  55.3  \\  
\hline
Our model & \textbf{77.7}  &  56.4 \\ 
\hline \hline
\end{tabular}
\end{table}
For all datasets, we employ the original conversation and knowledge base information only and drop the other labels e.g. slot filling labels. We take several metrics over all datasets to evaluate the performance on multiple dimensions. And to evaluate the context-aware memory networks, we also test the HMNs with only context-free memory on the dialog bAbI tasks.

\subsection{Evaluation Method}
 To compare with the original datasets baselines, we apply evaluation methods on each datasets the same as datasets' original papers described in \ref{sec:datasets}. 
\begin{itemize}
    \item Bilingual Evaluation Understudy (BLEU) \cite{DBLP:conf/acl/PapineniRWZ02}. BLEU has been widely employed in evaluating sequence generation including machine translation, text summarization, and dialogue systems. BLEU calculates the n-gram precision which is the fraction of n-grams in the candidate text which is present in any of the reference texts.
    \item F1 Score (F-measure): F1 evaluates the model's ability in terms of precision and recall, which is more comprehensive than just using precision or recall measure. We adopt F1 to evaluate if a model can extract information from a knowledge base precisely. 
    \item Per-response accuracy and Per-dialog accuracy. Per-response and Per-dialog accuracy count the percentage of responses that are correct. Any incorrect words will make a response or a dialogue negative. Accuracy shows if the model is able to learn the distribution of reproducing factual details.
\end{itemize}
\subsection{Baselines and Training Setup}
The hyper-parameter settings are adopted as the best practice settings for each training set following the Madotto's \cite{DBLP:conf/acl/FungWM18} and Manning's \cite{DBLP:conf/eacl/ManningE17} best experimental results on baselines SEQ2SEQ and Mem2Seq. Detailed models and their settings are as follows:
\begin{itemize}
    \item Sequence to sequence. For SEQ2SEQ, we adopt one layer LSTMs as encoder and decoder. For Key-Value Retrieval dataset, hidden size is placed at 512 and the dropout rate is 0.3. On dataset bAbI, the hidden size and dropout rate are 128 and 0.1 for task 3, 256 and 0.1 for task 4 and 5. Learning rates are set to 0.001 for bAbI and 0.0001 for DSTC 2 and Key-Value Retrieval dataset.
    \item SEQ2SEQ + Attention. We adopt the attention mechanism \cite{DBLP:conf/emnlp/LuongPM15} commonly used in neural machine translation. On dataset bAbI, hidden size and the dropout rate are 256 and 0.1 for task 3 and 4, 128 and 0.1 for task 5. For Key-Value Retrieval dataset, hidden size and dropout rate are 512 and 0.3. On the DSTC 2 task, hidden size is set to 353 and word embedding size is 300 (same with original work).
    \item Mem2Seq. Except 128 in task 3, hidden size in other tasks is 256. The dropout rate is set to 0.2 in task 3, 4 and Key-Value Retrieval dataset, 0.1 in task 5 and DSTC 2 dataset. We adopt three hops in DSTC 2 and Key-Value Retrieval dataset.
    \item HMNs with context-free only (HMNs-CFO). To test the performance of context-aware memory, we apply other context-free to encode dialogue history instead of context-aware memory in HMNs. All the other structure and parameter settings are the same as HMNs in this model.
\end{itemize}

All models are tested with various hyper-parameter settings to get their best performance, e.g. hidden size selected from [64, 128, 256, 512]. Note that settings from datasets are also tested like SEQ2SEQ + Attention's hidden size is 353 on Key-Value Retrieval dataset.

During the training, all experiments employ the teacher-forcing scheme, feeding the gold target of last time or highest probability word into decoder with probability 50\%. We also randomly mask input with UNK according to the dropout rate.
\begin{table*}[h]
\centering
\caption{A generated example from Key-Value Retrieval dataset with correct knowledge entities in \textbf{bold}. Given the knowledge base and user's request, we list the generated responses of three models and the gold answers. This example is randomly selected from all generated sentences and we only show tuples been used by models.} 
\label{fig:generated_examples}
\begin{tabular}{p{2.5cm}|p{12cm}}
\hline
\multicolumn{2}{c}{Dataset} \\ 
\hline
\multirow{8}{*}{Knowledge Base} & ...  \\
                 & valero poi\_type gas\_station \\
                 & valero distance 2\_miles \\
                 & valero address 200\_alester\_ave \\
                 & 1\_miles moderate\_traffic parking\_garage poi palo\_alto\_garage\_r \\
                 &  ...  \\
\hline
 User & address to the \textbf{gas\_station}  \\
\hline
 Gold  & \textbf{valero} is located at \textbf{200\_alester\_ave} \\
\hline
 \multicolumn{2}{c}{ Generated Sentence} \\
\hline
  SEQ2SEQ+Attn. & the closest \textbf{gas\_station} is located at \textbf{200\_alester\_ave} 7\_miles away would you like directions there \\
 Mem2Seq   & there is a \textbf{valero} 1\_miles away \\
 HMNs & there is a \textbf{gas\_station} located \textbf{2\_miles} away at \textbf{200\_alester\_ave}\\
\hline
\end{tabular}
\end{table*}
\subsection{HMNs Training Settings}
We test the hidden size in [64, 128, 256] and set dropout rate in [0.1, 0.2]. Learning rate is initiated with 0.001 and training batch is set to 64. The metrics results are coming from the best result settings for each dataset. We select hidden size and dropout rate at (256, 0.1) on bAbI task 3 and task 5, (256,0.2) on task 4. On the DSTC 2 task, we set hidden size and dropout rate at (128, 0.1). For Key-Value Retrieval dataset, the setting is hidden size 256 and dropout rate 0.1. Except for bAbI tasks' 1 layer, all HMNs and Mem2Seq tasks employ 3 layer memories. 

\begin{figure*}
\centering
\subfigure[Loss changes on bAbI T5]{
\begin{minipage}{0.29\linewidth}
\vspace{0.08in}
\includegraphics[width=1.0\linewidth]{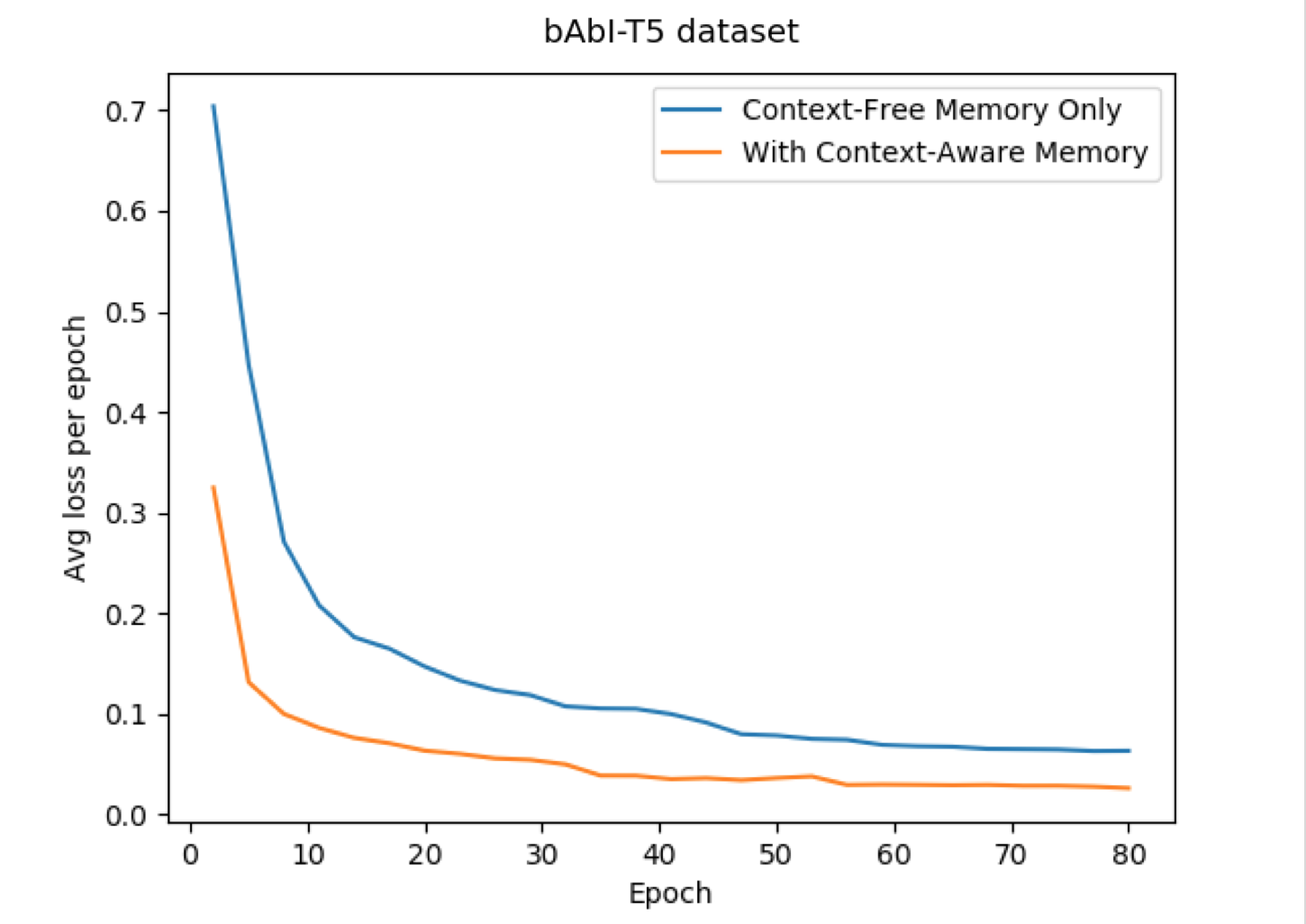}
\end{minipage}}
\subfigure[Loss by epochs on Key-Value Retrieval dataset]{
\begin{minipage}{0.3\linewidth}
\includegraphics[width=1.0\linewidth]{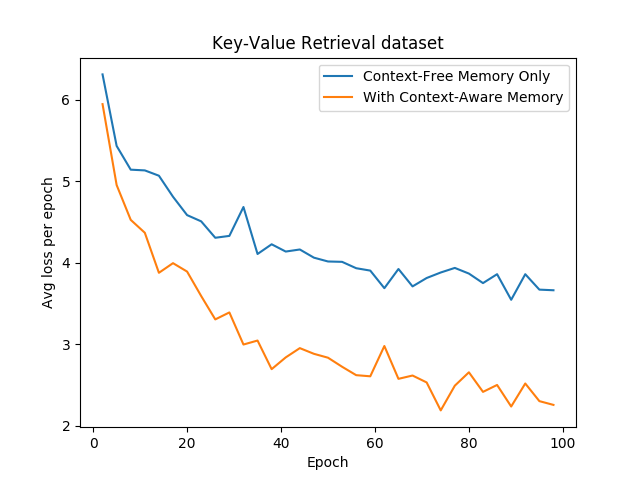}
\end{minipage}}
\subfigure[Loss by epochs on the DSTC2 dataset ]{  
\begin{minipage}{0.3\linewidth}
\includegraphics[width=1.0\linewidth]{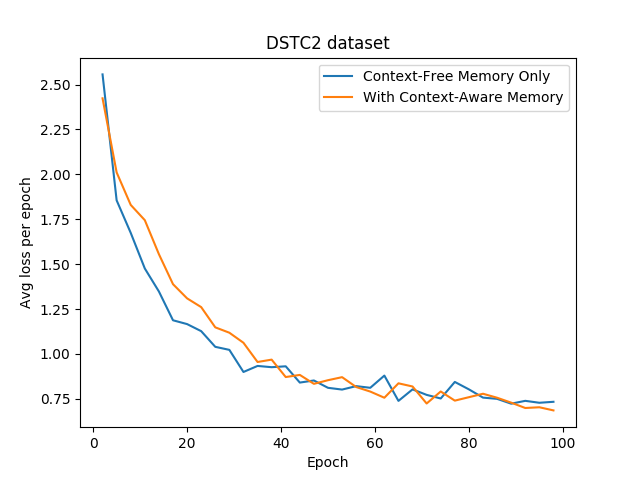}
\end{minipage}}
\caption{The total loss by epochs on three different datasets in 100 epochs. One line with context-aware memory and the other with only context-free memory }
\label{fig:loss_two_memories}
\end{figure*}
\section{Results and Discussion}
\subsection{Results and Analysis}
The best results of the baselines and HMNs are gathered into tables and figures.
Table \ref{table:results_kvr} show the result of models on Key-Value Retrieval dataset. Except for F1 scores on Calendar Scheduling, HMNs get significantly better results on all benchmarks comparing the state-of-the-art models. HMMs' BLEU score is even higher than human results which are reported in \cite{DBLP:conf/sigdial/EricKCM17}. Results show our model's outstanding performance in generating a fluent and accurate response in most tasks. 

Examples generated by our approach and baselines are given in Table \ref{fig:generated_examples}. These two examples are randomly selected from all generated sentences. Comparing the generated sentences by humans, although entities and sentences are different with gold answer in example one, our approach is able to produce more fluent and accurate sentences. However, the result on task weather forecasting neither HMNs and Mem2Seq can outperform SEQ2SEQ. We will discuss it in the next section. 

Table \ref{table:results_dstc2} shows our model gets the best F1 score on dataset DSTC 2, while SEQ2SEQ with attention gets the best BLEU result.

Table \ref{table:results_babi} shows results of models on bAbI tasks. HMNs and Mem2Seq adopt one hop attention only and note that all results are the best performance of each model in 100 epochs. HMNs achieved the best results on most tasks except T5. HMNs-CFO also outperforms the other models. This demonstrates that both training multiple distributions over heterogeneous information and employment of context-aware memory benefit the end-to-end dialogue system. The improvements in per-dialogue accuracy on out-of-vocabulary tests are even more significant. Figure \ref{fig:loss_two_memories} shows the changes of HMNs and HMNs-CFO's total loss across time. HMNs learns significantly faster.

Though automatic metrics cannot really examine human beings' diversified expression, existing dialogue systems aim at generating sentence by learning the patterns of training data, so we believe BLEU is still a metric of great concern in comparing the similar models' ability in learning the sentence patterns. Though human results show end-to-end machines have still a long way to go (60.7 to 43.1). Compared to other models, HMNs significantly improves performance in retrieving correct knowledge entities.

\subsection{Discussion}
\subsubsection{Context-Aware Memory}
To show whether context-aware memory benefits conversation learning, on bAbI tasks, we also tested HMNs-CFO memory only. From Table \ref{table:results_babi}, we observe that HMNs-CFO is significantly better than original Mem2Seq as well as SEQ2SEQ + attention in several results and only loses slightly on task 4 (89.3 to 90.5). One reason is that one memory is difficult to learn best distribution over different sources. Respectively encoding sequential dialogue history and grounding knowledge can learn two better distributions than one general but not best distribution. This also indicates that using the query vector generated by history memory to retrieve information in knowledge base memory is reasonable.

As the HMNs model get the best results in all tasks except one, in addition the results of training speed of HMNs and HMNs-CFO (Figure \ref{fig:loss_two_memories}), the context-aware memory is clearly to learn representation of the dialogue history much better and faster and also demonstrates that the importance of incorporating context information for dialogue systems. 
HMNs outperform the HMNs-CFO not only on BLEUs but also entity F1 on most tasks, showing building a good representation of dialogue history benefits knowledge reasoning, and help to improve the context-free memory by issuing a good query vector.

From above all, we can conclude that both stacked memory networks architecture and using context-aware memory to load sequential information can improve the performance of retrieving knowledge and generating sentences.

\subsubsection{Shortcomings}
From the results in Table \ref{table:results_kvr}, we note that HMNs and Mem2Seq failed on weather forecasting task. We analysed the average knowledge pairs of weather forecasting tasks and find it near three times the knowledge pairs of the other two tasks. Then we carried out another experiment that first narrows the KB candidates by performing a matching preprocessing operation, and the Weather Ent. F1 result of our method will climb to more than 48 which is the best. This may indicates that this kind of memory networks may have difficulties in handling large scale knowledge base. So perform a matching operation to narrow the candidate knowledge space is critical in a real-world large scale knowledge base.

And in this paper, we only show sequential data and knowledge triples data. For more types of information to integrate, model needs to add other memory networks, e.g. graph neural networks augmented memory networks \cite{DBLP:conf/ijcai/ZhouYHZXZ18} for graph structured data.

\section{Conclusion}
In this paper, we propose a model that is able to incorporate heterogeneous information in an end-to-end dialogue system. The model applies Heterogeneous Memory Networks (HMNs) to model sequential history and structured database. 
Results on several datasets show model can significantly improve the performance of generating the response. Our proposed context-aware memory networks show outstanding performance in learning the distribution over dialogue history and retrieving knowledge.
We present the possibility of efficiently using various structured data in end-to-end task-oriented dialogue without any extra labeling and module training.  

\section*{Acknowledgement}
We thank the anonymous reviewers for their insightful comments on this paper. This work was supported by the NSFC (No.61402403), DAMO Academy (Alibaba Group), Alibaba-Zhejiang University Joint Institute of Frontier Technologies, Chinese Knowledge Center for Engineering Sciences and Technology, and the Fundamental Research Funds for the Central Universities.

\appendix

\bibliography{emnlp-ijcnlp-2019.bib}
\bibliographystyle{acl_natbib}

\end{document}